\documentclass[10pt,twocolumn,letterpaper]{article}

\usepackage{cvpr}              

\usepackage{graphicx}
\usepackage{multirow}
\usepackage[pagebackref,breaklinks,colorlinks]{hyperref}

\usepackage{amsmath}
\usepackage{amssymb}
\usepackage{booktabs}
\usepackage{enumitem}
\usepackage{graphicx}
\usepackage{wrapfig}
\usepackage{times}
\usepackage{soul}
\usepackage{lipsum}
\usepackage[accsupp]{axessibility}  

\usepackage[capitalize]{cleveref}
\crefname{section}{Sec.}{Secs.}
\Crefname{section}{Section}{Sections}
\Crefname{table}{Table}{Tables}
\crefname{table}{Tab.}{Tabs.}

\def\cvprPaperID{3} 
\def\confName{CVPR}
\def\confYear{2022}

\begin{document}

\title{Multimodal Transformer for Nursing Activity Recognition}

\author{Momal Ijaz$^{1}$, Renato Diaz$^{1}$, Chen Chen$^{1,2}$\\
$^1$Department of Computer Science, University of Central Florida, USA\\
$^2$Center for Research in Computer Vision, University of Central Florida, USA\\
{\tt\small \{im.momil, diazrenato2001\}@knights.ucf.edu, chen.chen@crcv.ucf.edu}}

\maketitle

\begin{abstract}
    In an aging population, elderly patient safety is a primary concern at hospitals and nursing homes, which demands for increased nurse care. By performing nurse activity recognition, we can not only make sure that all patients get an equal desired care, but it can also free nurses from manual documentation of activities they perform, leading to a fair and safe place of care for the elderly. 
    \par In this work, we present a multimodal transformer-based network, which extracts features from skeletal joints and acceleration data, and fuses them to perform nurse activity recognition. Our method achieves state-of-the-art performance of 81.8\% accuracy on the benchmark dataset available for nurse activity recognition from the Nurse Care Activity Recognition Challenge. We perform ablation studies to show that our fusion model is better than single modality transformer variants (using only acceleration or skeleton joints data). 
    \par Our solution also outperforms state-of-the-art ST-GCN, GRU and other classical hand-crafted-feature-based classifier solutions by a margin of 1.6\%, on the NCRC dataset. Code is available at \url{https://github.com/Momilijaz96/MMT_for_NCRC}.
    
\end{abstract}
\section{Introduction}
\label{sec:intro}

    Elderly care and safety is a primary concern at health care centers, which highly demands for increased nurse care. Performing nurse activity recognition is an important task as it can aid in the process of monitoring the health care plans compliance for each patient and frees up nurses from the task of manual reporting and documentation. Activities performed by nurses tend to be more complex and longer than straightforward actions or gestures available in benchmark data sets like walking, running, eating, sleeping, and waving hello \cite{liu2020ntu, shahroudy2016ntu}. 
    
    Human activity recognition is a widely researched area in computer vision as it has applications in human computer interaction or video understanding, etc. \cite{Wang2016TemporalSN, 8099985, chen2014home}. Over the past few years, skeleton-based action recognition has gained popularity because of it's good estimate on human body's dynamic movements and is also more robust to illumination variations and background noises \cite{7486569, Song2017AnES}.
  
     Skeletal action recognition has been previously performed using hand crafted features \cite{6909476} or manually structuring data as a pseudo image and passing it to a Convolutional Neural Network (CNN) \cite{7486569}, or as a sequence of coordinates vectors which are fed to a Recurrent Neural Network (RNN) \cite{10.1007/978-3-319-46487-9_50, XieLZ0HL18}. Other works have explored the benefits of using divided space-time feature extraction by using a Spatio-Temporal Graph Convolutional Network (ST-GCN), which models the spatial configuration and temporal dynamics of skeletons  \cite{Yan2018SpatialTG}. Recently, researchers have been trying to import the capabilities of transformers \cite{Vaswani2017AttentionIA} from Natural Language Processing (NLP) to vision domain. Among all other variants, recent Vision Transformers \cite{Dosovitskiy2021AnII} stood out, as they showed a convolution-free transformer network can show comparable performance to CNNs in vision tasks. Similar studies have been done by researchers for coming up with a pure transformer architecture for skeletal action recognition as well \cite{10.1007/978-3-030-68796-0_50, Wang2021IIPTransformerIT}.
     
     In addition to skeletal poses, acceleration signal has been used in quite a few works\cite{10.1007/978-3-540-24646-6_1, 7489944, BAYAT2014450} for performing action recognition and acceleration has proven to be quite effective for the task. The Nurse Care Activity Recognition Challenge (NCRC) dataset\cite{inoue2019nurse} comprises of data from acceleration sensors, location sensors and skeletal joints. Previous works on the dataset used different combinations of these modalities with  hand crafted features, using simple classification algorithms like KNN or Random-Forests \cite{10.1145/3341162.3344859, lago2019nurse}. Other advanced works on the dataset used ST-GCNs \cite{10.1145/3341162.3345581} and Gated Recurrent Units (GRUs) \cite{10.1145/3341162.3344848}. All these works used different combinations of data modalities (i.e. skeletal joints, location, and acceleration), but none of the works explored fusion of the two strongest signals for action recognition, i.e. acceleration \& skeletal joints. 
    
    In this paper, we present a multimodal transformer network that fuses acceleration and spatio-temporal skeletal features to perform activity recognition on the NCRC dataset. The main contributions of our work are summarized as follows:
    
    \begin{itemize}
    \item We show that fusing acceleration signal and skeletal joints data leads to improved performance for action recognition, as compared to using single modality. Additionally, we present pure transformer-based single modality networks for skeletal joints and acceleration data, and an efficient dual modality network for both signals. Our dual modality transformer, using both acceleration and skeletal joints data, outperforms single modality networks by 5.2\%.
    \item We present a novel attention-based fusion technique for fusing spatio-temporal skeletal features with acceleration features, for exploiting correlations between acceleration and skeletal joints; to develop better semantic understanding of actions being performed. Our fusion method outperforms simple fusion baseline, by 6.8\%.
    \item Our proposed dual modality transformer outperforms state-of-the-art GRUs, ST-GCNs, and handcrafted feature-based classifiers, like KNNs, and achieves the highest performance on the NCRC dataset of 81.8\%.
    \end{itemize}

\section{Related Work}

\begin{figure}
\centering
    \includegraphics[width=0.95\linewidth]{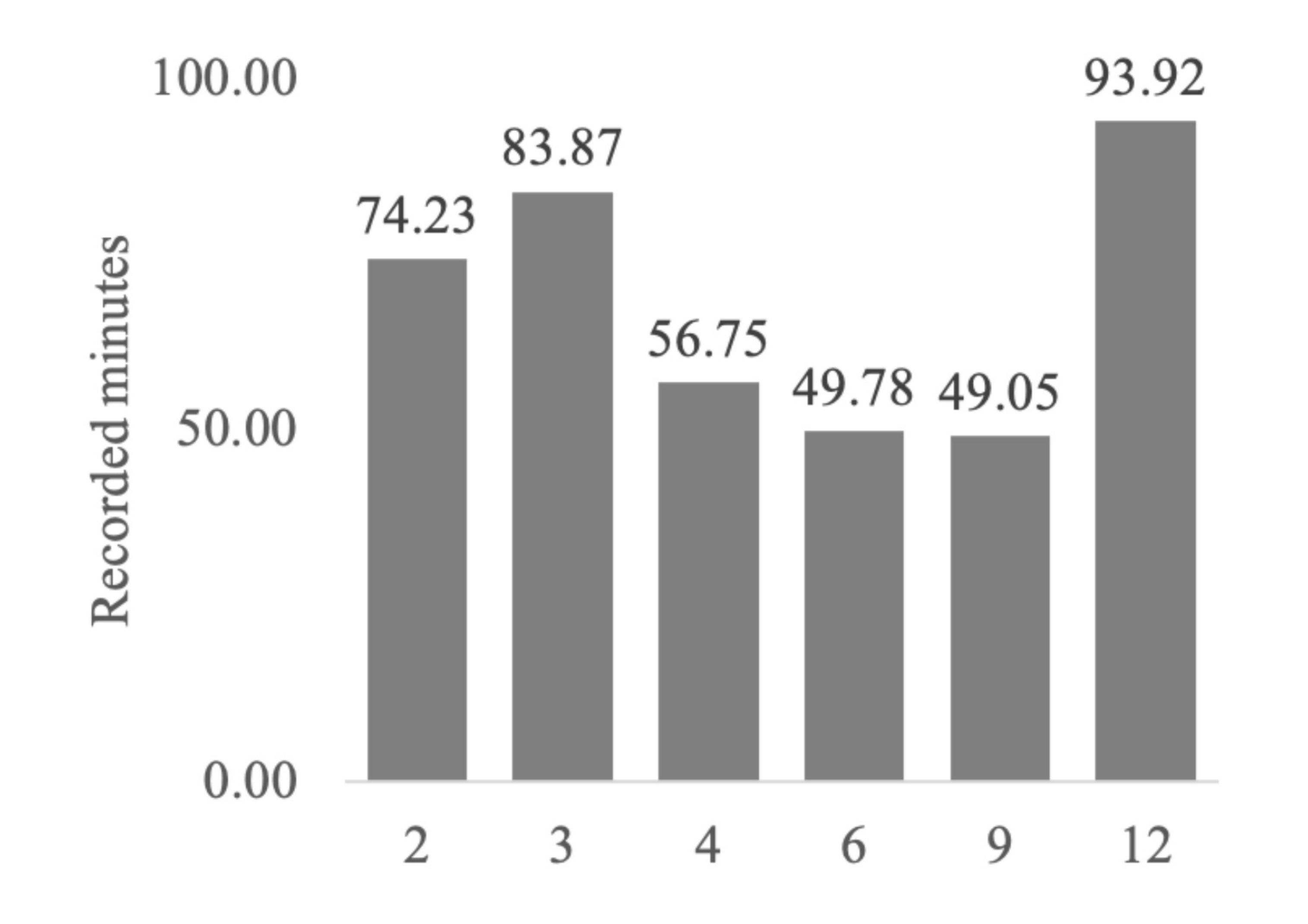} %
    \caption{Individual recorded time for each activity. Activity2: Vital Signs Measurement, Activity3: Blood collection, Activity4: Blood glucose measurement, Activity6: Indwelling drip retention and connection, Activity 9: Oral care, Activity 12: Diaper exchange and cleaning of area. }%
    \label{fig:example}%
\end{figure}

\begin{figure*}[h]
\centering
    \includegraphics[width=17cm,height=8cm]{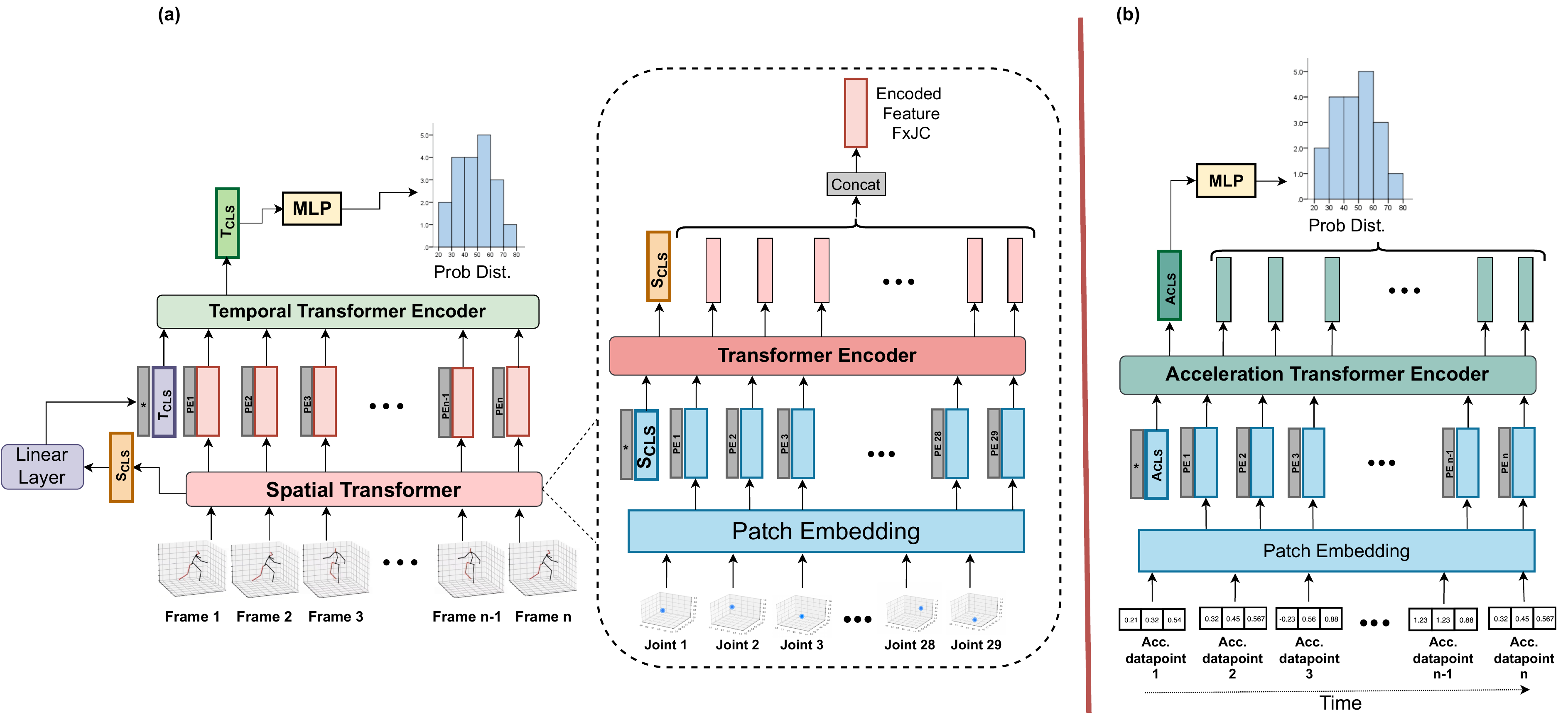} %
    \caption{\textbf{Single Modality Models:} Single modality models use only acceleration or skeletal joints data. \textbf{(a) Spatio-Temporal Skeleton Model: } The skeletal model comprises of two transformer blocks: spatial and temporal encoders, for computing spatial and temporal features from skeletal joints of  given action sample. \textbf{(b) Acceleration Model:} The acceleration model has one transformer block, which computes correlation across acceleration data-points for a given action sample.}%
    \label{fig:example}%
\end{figure*}

\par \textbf{Skeletal Action Recognition:}
Skeletal action recognition has lately been a preferred method for performing action recognition since it is more robust to illumination variations and other background noises. Older methods such as manual extraction of hand crafted features\cite{6909476}, crafting pseudo images out of the skeletal poses and feeding them into CNNs\cite{7486569}, or other RNN based methods\cite{10.1007/978-3-319-46487-9_50} have become outdated after the huge success of graph-based methods\cite{Yan2018SpatialTG}. Yan et al. introduced ST-GCNs which are able to map spatial correlations and temporal changes of a human skeleton for performing action recognition. GCN-based approaches \cite{Cheng2020SkeletonBasedAR, shi2018non} use topographical features of skeleton to extract and combine spatial skeletal features and temporal dynamics. In contrast to ST-GCN-based methods, transformers can directly learn correlations between joints in a frame and complete skeletal poses across frames.  Using a divided space-time attention mechanism, researchers have shown different variants of transformer architecture for performing skeletal action recognition. In \cite{10.1007/978-3-030-68796-0_50}, authors use pure transformer architectures to map correlation between joints in one frame and  across frames, using two different transformers for spatial and temporal feature extraction. In \cite{Wang2021IIPTransformerIT}, authors group joints into parts. They use a single transformer encoder block for computing spatial and temporal features of the skeletal joints data. Their proposed method involves computing correlations between joints in one part, across parts in one frame and across frames for same part, using a modified intra-inter part attention mechanism. 

\begin{figure}
\centering
    \includegraphics[width=8.5cm, height=6cm]{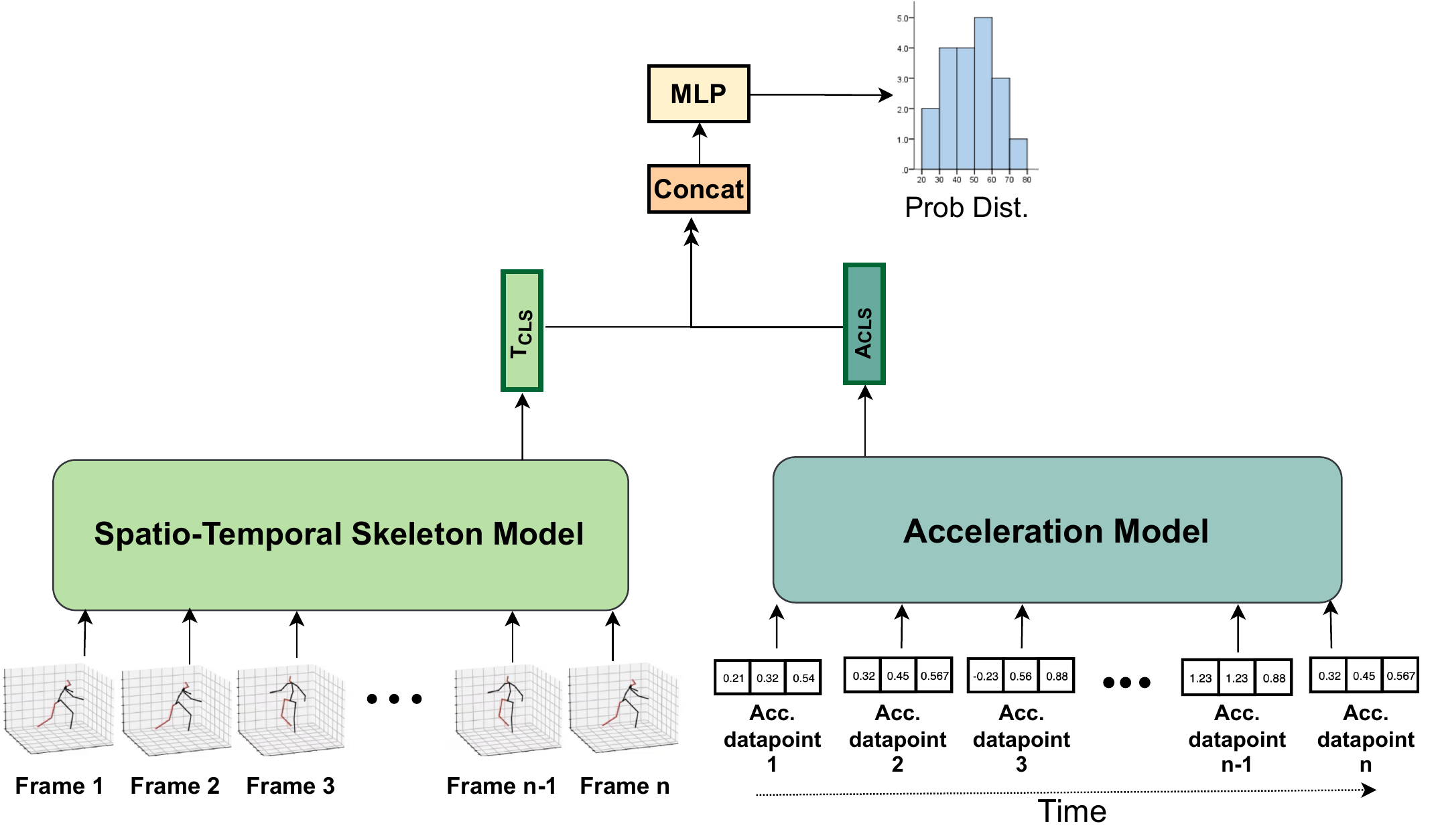} %
    \caption{ \textbf{Multi-Modal Transformer - Simple Fusion}. Multi-modal transformer uses both modalities, acceleration and skeletal joints. Skeletal features are extracted by single modality skeleton model, whereas the acceleration features are extracted by single modality acceleration model, and skeletal and acceleration features are added to perform classification.}%
    \label{fig:example}%
\end{figure}

\begin{figure*}[h]
\centering
    \includegraphics[width=\textwidth,height=7cm]{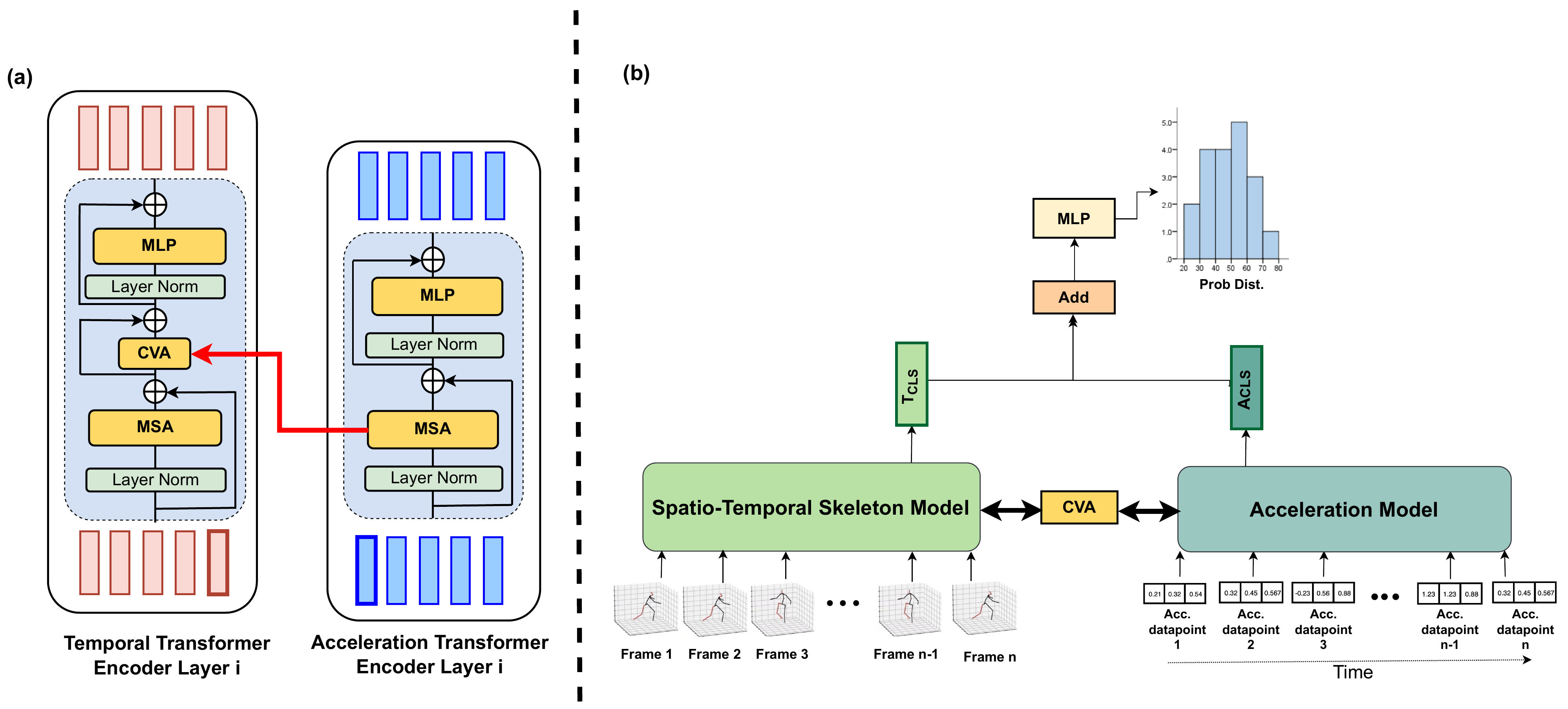} %
    \caption{\textbf{(a) CrossView Fusion}. Cross View fusion is performed between corresponding encoder layers of acceleration block from acceleration model and temporal transformer block from spatio-temporal skeleton model.
    \textbf{(b) Multi-Modal Transformer - Cross View Fusion}. Cross view fusion-based multimodal transformer is exactly similar to simple fusion transformer (Figure-1), except for an added cross view fusion mechanism between acceleration and skeletal joints branch.}%
    \label{fig:example}%
\end{figure*}

\par \textbf{Acceleration-based Action Recognition:}
 Acceleration signal has proven to be useful for performing action recognition. Most earlier works used hand crafted statistical features from acceleration signal with simple classifiers like Support Vector Machines (SVMs) \cite{10.1007/978-3-540-24646-6_1, 7489944, BAYAT2014450}. However, novel deep-learning-based techniques have outperformed the classical approaches significantly. In \cite{s20226424}, the authors use a 3-layered CNN, followed by a Long Short Term Memory (LSTM) block for performing activity classification, and show that their method is better than SVM trained on similar features. In  \cite{7379395}, the authors present a pure CNN architecture which exceeds classical feature extraction pipelines. The proposed CNN architecture has a modified convolutional kernel to adapt to the triaxial acceleration signal. However, as mentioned in \cite{Tong2020AreAF}, accelerometer-based activity recognition is considered a dead end as the sensor offers limited information. We extend this idea and explore the effectiveness of fusing skeletal joints data with acceleration signal for activity recognition. 
 
\par \textbf{Nurse Care Activity Recognition:} Best performance on the NCRC dataset was obtained by using handcrafted features extracted from skeletal data and location sensor with an ensemble of  K-Nearest Neighbors (KNNs) models\cite{10.1145/3341162.3344859}. Other works explored using RandomForests\cite{lago2019nurse} with just acceleration signal. Among deep-learning approaches, baseline set up by competition organizers comprised of a CNN backbone which used all data modalities\cite{lago2019nurse}. Other works used ST-GCNs\cite{10.1145/3341162.3345581} with skeletal joints data, and GRUs \cite{10.1145/3341162.3344848} with skeletal joints and location sensor data. We explore the fusion of acceleration signals with skeletal joints with a transformer-based network.

\par \textbf{Transformers:} Researchers have been trying to explore the capabilities of revolutionary transformers\cite{Vaswani2017AttentionIA} in vision domain, to see if these are a strong competitor against the widely used CNNs. Different variants of transformers have shown promising results for various vision tasks including but not limited to video and image classification, semantic segmentation, and object localization \cite{Dosovitskiy2021AnII, Liu2021SwinTH, chen2021crossvit}.

\par \textbf{Transformer-based Fusion Strategies:}
Various mechanisms have been studied for the exchange or fusion of information between two transformer blocks. In \cite{ramazanova2022owl}, the authors fuse audio and visual signal using a simple early fusion technique to perform video classification. In CrossViT\cite{chen2021crossvit}, authors utilize tokens to exchange information between two transformer blocks that process images of two different resolutions. The fusion technique used in CrossViT is based on cross-attention, in which  tokens of one branch attend to encoded features of other branches for sharing information. As for video classification, in the state-of-the-art Multi-view Transformer \cite{Yan2022MultiviewTF}, the authors compare three different fusion techniques to exchange information between different resolution of video tokens. They found cross view fusion to be the best fusion approach, in which tokens of larger resolution attend to tokens of smaller resolution, at selected layers in a transformer block. Our proposed fusion approach, for combining information from skeletal joints and acceleration signal, is majorly inspired from this technique.

\section{Dataset}

The dataset used for the study along with it's posed challenges is as follows.
\subsection{Description}
For this experimentation, we are using the dataset from the Nurse Care Activity Recognition Challenge\cite{inoue2019nurse}. In this particular dataset, 6 different activities have been recorded by 8 nurses working in a controlled, monitored environment. These activities are as under:

\begin{itemize}[itemsep=0.8mm, parsep=0pt]
\item Vital signs measurement 
\item Blood collection 
\item Blood glucose measurement 
\item Indwelling drip retention and connection 
\item Oral care 
\item Diaper exchange and cleaning of area
\end{itemize}

The dataset comes with a testing and training split. The training set contains all the aforementioned activities performed by 6 subjects, which sums up to a total of 282 action samples. Whereas the testing set contains actions performed by 2 different subjects, summing up to a total of 116 samples. Motion-capture cameras, accelerometer chips, and location sensors are used to record each action. The motion capture camera captures 29, 3D, joint locations at 100 Hz frequency. The accelerometer chip captures the acceleration of the subject along x, y and z axes at 4Hz frequency. The location sensor captures the x and y location coordinates of the subject and changes in air pressure at 20Hz frequency. Acceleration data is captured using a sensor, placed in upright position in the right pocket of the subject, whereas skeletal data is collected using IR-based motion capture cameras. Figure 1 shows the recorded minutes of each activity.  

\subsection{Challenges Posed by the Dataset}
\par This dataset is the only one of it's kind that fulfills all of our requirements, i.e. being designed for nurse care activity recognition. However, there are a few challenges posed by this dataset, detailed below.
\begin{itemize}[itemsep=1mm, parsep=0pt]
\item The sampling rate of sensors are widely different, Figure 5. Skeletal data was recorded at 100Hz, giving 6000 skeletal poses per action, whereas acceleration data was recorded only at 4Hz, giving roughly 150 data points per action. That's 0.25 data points for acceleration and 100 for skeletal joints in a second.  
\item Acceleration data was very noisy. It had null values and was entirely missing for 2 action samples performed by subject 2.
\item Overall size of the dataset was very small, there were just 282 training and 216 testing samples available, which proved to be a main road-block in training data hungry transformer-based architecture.
\item Skeletal data also had entirely missing joints for some action samples and was noisy as well. 
\end{itemize}

\section{Method}

\par In this study, we explore the fusion of acceleration features with spatio-temporal skeletal features for performing nurse activity recognition. The acceleration data and skeletal joints data are convenient and economical to collect. For acceleration, we have lightweight accelerometer chips or smartphones. For skeletal data, motion capture cameras like Kinetics \cite{kinetics} and RealSense \cite{Keselman2017IntelRS} can do a pretty decent job even in a hustling nursing environment. 

\subsection{Single Modality Transformers}
We present 2 single modality transformer models, which are trained only on acceleration or skeletal joints data. Each single modality transformer model comprises of a class token, like ViT\cite{Dosovitskiy2021AnII}, which is used for performing final classification. 
\subsubsection{Spatio-Temporal Skeleton Model}  
\par Figure 2(a) demonstrates the architecture of a spatio-temporal model, which is mainly inspired from the PoseFormer\cite{zheng20213d} network. The spatio-temporal skeletal model is a single modality transformer that performs action recognition using only skeletal joints data. 

\par The model comprises of two transformer blocks, spatial and temporal. Spatial transformer takes each frame as an input and computes correlation between 29 individual 3D joint points in a frame. We pass each 3D joint coordinate through a linear patch embedding layer and add position encoding before passing it to a standard transformer encoder block. We also append a spatial CLS token \(S_{cls}\) to the inputs. The spatial transformer outputs feature vectors for each joint, which are concatenated together. This concatenated feature vector is a representation of each frame computed by the spatial transformer. This process is repeated for all frames in the video sample and finally we pass all encoded frames to the temporal transformer block for computing temporal correlation across frames. In the spatial transformer, each token is a joint, whereas for temporal transformer each token is a feature vector representing one frame.

\par The spatial CLS, \(S_{cls}\),  token is passed through a linear layer to project it up to the temporal embedding dimension. This token with temporal embedding dimension is called \(T_{cls}\) and is passed to the temporal transformer encoder along with other encoded frames. \(T_{cls}\) is used for final classification and hence is passed through a simple linear MLP classification head and gives probability distribution of labels.

\subsubsection{Acceleration Model} 
\par This model, Figure 2(b), attempts to perform action recognition using just the acceleration of the performer. Each acceleration data point comprises of acceleration value, recorded every 4 seconds, along the x, y, and z dimensions. We interpolate the acceleration signal using simple linear interpolation. Next, we denoise it using a moving average window of size 40, and fill in the samples with missing acceleration data.
\par The acceleration-only model is similar to the spatial transformer model, although here each token is an acceleration data point (3D vector). We encode each data point using a linear embedding layer, append position encodings, and an acceleration CLS token, \(A_{cls}\), with inputs, which is passed through acceleration transformer block. The output is the encoded feature vector \(A_{cls}\), which is passed through a MLP head for the prediction of the target class.

\begin{table}
  \centering
    \label{tab:table1}
    \resizebox{0.95\columnwidth}{!}{
    \begin{tabular}{|c|c|c|c|c|}
    \hline
      \textbf{Model} & \textbf{Learning Rate} & \textbf{Drop} & \textbf{Stoch. Drop} & \textbf{Attn. Drop}\\
      \hline
      Skeleton only & 0.02 & 0 & 0.2 & 0\\ \hline
      Acceleration only & 0.02 & 0 & 0.2 & 0\\ \hline
      Simple Fusion & 0.0025 & 0.05 & 0.2 & 0.05\\ \hline
      CrossView Fusion & 0.0025 & 0 & 0.2 & 0 \\ 
      \hline
    \end{tabular}}
  \caption{\textbf{Training Hyper-parameters}. Single Modality models performed well without strong regularization, whereas fusion models converged well with non-zero drop rates. }
\end{table}

\subsection{Multi-Modal Transformers}
\par We use the single modality transformer models to create 2 different dual modality transformer models, which utilize both acceleration and skeletal joints data. The first dual modality transformer is a simple feature baseline, which concatenates the respective class tokens from acceleration and skeletal joints branch to perform classification. The second dual modality transformer is similar to the first one with just the addition of cross view fusion mechanism. This mechanism allows for the exchange information between skeletal joints and the acceleration branch. 

\subsubsection{Simple Fusion} 
\par Figure 3 illustrates the simple fusion model, inspired from the early fusion technique presented in \cite{ramazanova2022owl}, for fusing visual and audio signal.
In the simple fusion model, we take the single modality spatio-temporal skeleton model and acceleration model. Skeleton model takes skeletal joints as input and computes spatio-temporal skeletal features and gives a temporal CLS token, \(T_{cls}\), as output. The acceleration model takes acceleration of the same action as input and outputs an acceleration CLS token \(A_{cls}\). We simply concatenate these two CLS tokens and pass them to the MLP classification head, which gives us the resultant class of action sample.

\subsubsection{CrossView Fusion} In CrossView fusion model, along with simple aggregation of the CLS tokens from both branches, we fuse information between acceleration and skeletal encoders. Particularly, the tokens of the temporal transformer block of the spatial-temporal skeleton model act as queries, and the tokens of acceleration encoder block act as key and value pairs. This fusion technique is majorly inspired from the CrossView attention presented in Multi-view transformer\cite{Yan2022MultiviewTF} paper, for fusing information from multi resolution input patches. CrossView fusion model allows the temporal skeletal joint features to attend to acceleration features for developing a better understanding of the action being performed. 
For CrossView fusion, we keep the embedding dimension and depth of both acceleration and temporal encoders similar, which eliminates the need to project up or down the tokens before passing to other branch. CrossView fusion introduces an additional cross attention operation after multi-headed self attention (MSA) mechanism in each layer of the temporal skeletal encoder. This block attends to MSA encoded acceleration tokens from the corresponding layer.
Mathematically, each \(i^{th}\) layer in temporal skeletal encoder attends to the \(i^{th}\) layer of acceleration encoder, as shown in the equation below.
\[z^{temporal_i} = CVA(z^{temporal_i},z^{acc_i})\]
\[CVA(x,y) = Softmax(\frac{W^{Q}xW^{K}y^{T}}{\sqrt d_{k}})W^{V}y\]
Here, CVA stands for CrossView Attention, \(z^{temporal}\) is temporal skeletal encoder tokens, and \(z^{acc}\) are acceleration encoder tokens, with \(W^{Q}\), \(W^{K}\) and \(W^{V}\) as the weights of CVA block for computing query, key and value representations.

\begin{table*}[h]
\centering
 \scalebox{0.75}{
\begin{tabular}{|c|c|c|}
\hline
\textbf{Sensors Used}           & \textbf{Method}  & \textbf{Validation Accuracy (\%)} \\ \hline

Motion Capture and Location     & KNN                                      & 80.2                              \\ \hline
Motion Capture                  & ST-GCN                                      & 64.6                              \\ \hline
All modalities                  & CNN                                      & 46.5                              \\ \hline
Acceleration                    & Random Forest                            & 43.1                              \\ \hline
Motion Capture and Location     & GRU                                           & 29.3                              \\ \hline
\textbf{Acceleration and Motion Capture (Our Approach)} & \textbf{Transformers}                               & \textbf{81.8}                     \\ \hline
\end{tabular}}
\caption{\textbf{Comparison with state-of-the-art:} Comparing our approach with other modalities and methods on the NCRC dataset. Our proposed method, fusion of acceleration and skeletal joints using transformer-based method outperforms all other modalities and methods.}
\end{table*}

\begin{table*}[h]
\centering
\resizebox{1.0\linewidth}{!}{
\begin{tabular}{|c|ccccc|ccccc|}
\hline
{\textbf{Activity-ID}} & \multicolumn{5}{c|}{\textbf{Accuracy(\%)}}                                                                                                                                & \multicolumn{5}{c|}{\textbf{F1-Score}}                                                                                                                                  \\ \cline{2-11} 
                                    & \multicolumn{1}{c|}{\textbf{Our Approach}} & \multicolumn{1}{c|}{\textbf{ST-GCN}} & \multicolumn{1}{c|}{\textbf{RF}} & \multicolumn{1}{c|}{\textbf{DTT}} & \textbf{KNN}   & \multicolumn{1}{c|}{\textbf{Our Approach}} & \multicolumn{1}{c|}{\textbf{ST-GCN}} & \multicolumn{1}{c|}{\textbf{RF}} & \multicolumn{1}{c|}{\textbf{DTT}} & \textbf{KNN} \\ \hline
2                                  & \multicolumn{1}{c|}{\textbf{80.00}}         & \multicolumn{1}{c|}{65.10}            & \multicolumn{1}{c|}{4.17}        & \multicolumn{1}{c|}{52.08}        & 47.92          & \multicolumn{1}{c|}{\textbf{80.00}}         & \multicolumn{1}{c|}{63.30}            & \multicolumn{1}{c|}{5.97}        & \multicolumn{1}{c|}{54.95}        & 61.33        \\ \hline
3                                  & \multicolumn{1}{c|}{75.0}                  & \multicolumn{1}{c|}{54.5}            & \multicolumn{1}{c|}{68.25}       & \multicolumn{1}{c|}{73.02}        & \textbf{95.24} & \multicolumn{1}{c|}{65.5}         & \multicolumn{1}{c|}{48.9}            & \multicolumn{1}{c|}{67.19}       & \multicolumn{1}{c|}{69.70}        & \textbf{76.43}        \\ \hline
4                                  & \multicolumn{1}{c|}{\textbf{71.0}}         & \multicolumn{1}{c|}{60.0}            & \multicolumn{1}{c|}{13.89}       & \multicolumn{1}{c|}{38.89}        & 22.22          & \multicolumn{1}{c|}{\textbf{72.8}}         & \multicolumn{1}{c|}{55.0}            & \multicolumn{1}{c|}{21.74}       & \multicolumn{1}{c|}{43.08}        & 34.78        \\ \hline
6                                  & \multicolumn{1}{c|}{\textbf{85.8}}          & \multicolumn{1}{c|}{62.2}            & \multicolumn{1}{c|}{84.38}       & \multicolumn{1}{c|}{34.38}        & 21.88          & \multicolumn{1}{c|}{\textbf{82.8}}         & \multicolumn{1}{c|}{65.3}            & \multicolumn{1}{c|}{67.50}       & \multicolumn{1}{c|}{37.93}        & 32.56        \\ \hline
9                                  & \multicolumn{1}{c|}{\textbf{71.5}}         & \multicolumn{1}{c|}{50.8}            & \multicolumn{1}{c|}{12.12}       & \multicolumn{1}{c|}{0.00}         & 57.58          & \multicolumn{1}{c|}{\textbf{77.0}}         & \multicolumn{1}{c|}{44.2}            & \multicolumn{1}{c|}{17.39}       & \multicolumn{1}{c|}{0.00}         & 67.86        \\ \hline
12                                  & \multicolumn{1}{c|}{\textbf{87.1}}         & \multicolumn{1}{c|}{49.0}            & \multicolumn{1}{c|}{90.77}       & \multicolumn{1}{c|}{47.69}        & 100            & \multicolumn{1}{c|}{\textbf{93.2}}         & \multicolumn{1}{c|}{40.5}            & \multicolumn{1}{c|}{63.10}       & \multicolumn{1}{c|}{40.79}        & 73.45        \\ \hline
\textbf{Class Wise Mean}                    & \multicolumn{1}{c|}{\textbf{78.2}}         & \multicolumn{1}{c|}{57.0}            & \multicolumn{1}{c|}{50.54}       & \multicolumn{1}{c|}{45.85}        & 65.70          & \multicolumn{1}{c|}{\textbf{78.6}}         & \multicolumn{1}{c|}{52.9}            & \multicolumn{1}{c|}{43.82}       & \multicolumn{1}{c|}{44.92}        & 61.61        \\ \hline
\end{tabular}}
\caption{\textbf{Activity wise performance comparison with state-of-the-art}. Our method outperforms all existing solutions in terms of Accuracy and F1 Score except for Class 3 which is blood collection. Overall, Class Wise mean of accuracy and F1-scores of our approach is 12.5\% better than state-of-the-art KNN based solution.}
\end{table*}

\section{Experiments and Results}

\subsection{Implementation Details}
The NCRC dataset has a limited number of samples, and training a transformer based network requires strong data augmentation or pre-training. Although, we were able to converge our transformer models without using any of the two, by using adaptive sharpness aware minimization\cite{Kwon2021ASAMAS} (ASAM). This technique has been tested out on ViT\cite{Chen2021WhenVT}, and authors in this work were able to make ViTs outperform ResNet without augmentation or pre-training using sharpness aware minimization. We used ASAM with neighborhood size of 0.5, to smooth out the loss function and avoid over-fitting. ASAM focuses on finding optimal neighborhoods for network parameters instead of optimal values, which ultimately leads to a much smoother loss function and better generalization.
\par Along with various other regularization techniques, we also used stochastic depth\cite{Huang2016DeepNW}, which is known to facilitate the convergence of deep transformers.\cite{Fan2020ReducingTD} We set the stochastic depth rate in the range of 0.1 - 0.2. Additionally, we also used drop outs and attention drop rates, and found them key factor for allowing our fusion models to converge and generalize well. However, single modality models performed well with drop rates set to zero. We used a batch size of 16, with SGD optimizer, a weight decay rate of 5e-4, and Cosine Annealing learning rate scheduler. The rest of the hyper-parameters for converging every model are given in Table 1.

\begin{table}[h]
  \begin{center}
    \label{tab:table1}
    \scalebox{0.8}{
    \begin{tabular}{|l|c|c|c|c|}
    \hline
      \textbf{Model}  & \textbf{Accuracy} & \textbf{F1-Score} & \textbf{Precision} & \textbf{Recall}\\
      \hline
      Skeleton Model  & 76.7 & 67.0 & 69.1 & 70.5
       \\ \hline
      Acceleration Model  & 45.6 & 10.9 & 9.3 & 14.9\\ \hline
      Simple Fusion & 75.0 & 71.6 & 75.6 & 72.3\\ \hline
      Cross-View Fusion  & \textbf{81.8} & \textbf{78.4} & \textbf{79.4} & \textbf{78.3}\\ \hline
    
    \end{tabular}}
  \end{center}
  \caption{\textbf{Single Modality vs. Dual Modality Performance Comparison}  Dual modality CrossView fusion model outperforms single modality and simple fusion models.}
\end{table}

\begin{table}[]
\scalebox{0.7}{
\begin{tabular}{|c|c|c|c|c|}
\hline
\textbf{Nurse ID} & \textbf{Precision (\%)} & \textbf{Recall (\%)} & \textbf{F1-score (\%)} & \textbf{Accuracy (\%)} \\ \hline
2                 & 59.2                    & 65.8                 & 59.9                   & 67.4                   \\ \hline
3                 & 68.3                    & 67.9                 & 66.9                   & 72.7                   \\ \hline
4                 & 68.7                    & 68.4                 & 66.9                   & 80.0                   \\ \hline
5                 & 72.6                    & 73.1                 & 70.3                   & 74.6                   \\ \hline
6                 & \textbf{95.4}           & \textbf{91.5}        & \textbf{92.5}          & \textbf{93.4}          \\ \hline
7                 & {\ul \textit{44.8}}     & {\ul \textit{60.9}}  & {\ul \textit{50.6}}    & {\ul \textit{61.1}}    \\ \hline
8                 & 79.2                    & 77.2                 & 78.0                   & 79.3                   \\ \hline
9                 & 82.7                    & 79.4                 & 76.9                   & 82.3                   \\ \hline
\end{tabular}}
\caption{\textbf{LOSOCV Performance of CrossView Fusion Model}. The best performing subject ID is 6 (in bold), whereas the worst performing subject ID is 7(in italic).}
\end{table}

\begin{figure}
\centering
    \includegraphics[width=0.95\linewidth]{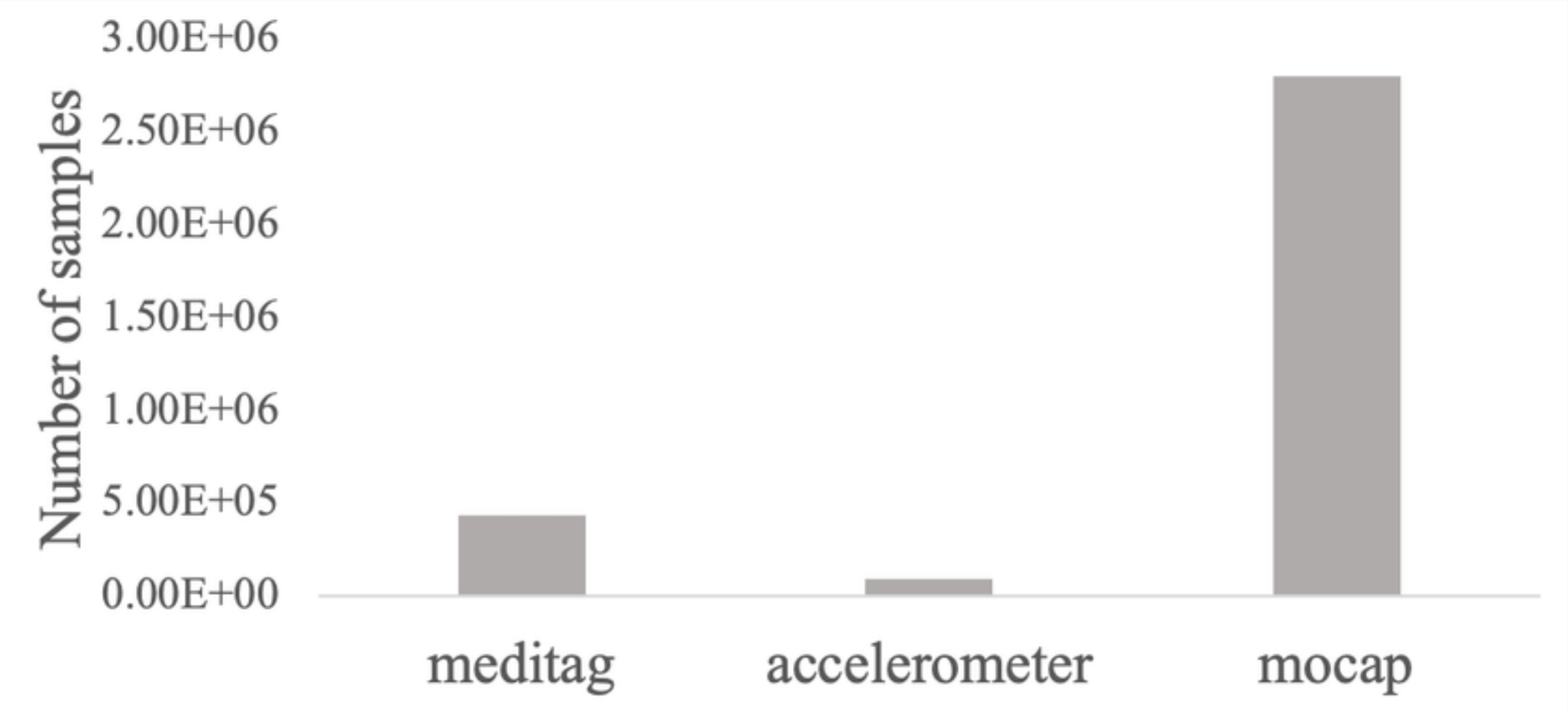} %
    \caption{Number of Observations by each sensor}%
    \label{fig:example}%
\end{figure}

\subsection{Comparison with State-of-the-art}
\par Earlier works that have also performed on this dataset are summarized in Table 2. The NCRC dataset comprises of three different modalities: acceleration, skeletal joints (Motion Capture), and location sensors. Among different combinations of these modalities, the fusion of acceleration with the skeletal joints performs the best and gives the highest validation accuracy. Reliance on just one signal like skeletal joints or acceleration does not provide adequate results. The second best performance is obtained by an ensemble of a KNN-based method \cite{10.1145/3341162.3344859}, which uses location and skeletal joints data with hand crafted features. We did not use the location signal in our method as the collection of location data in a nursing environment is relatively harder in the real world as compared to skeletal joints or acceleration data. Deep learning-based solutions like CNN\cite{lago2019nurse}, ST-GCN\cite{10.1145/3341162.3345581}  and GRU\cite{10.1145/3341162.3344848} perform poorly compared to our transformer-based solution due to the smaller size of the dataset. The usage of ASAM\cite{Kwon2021ASAMAS} clearly helps our model to avoid over-fitting compared to these solutions.

\par In terms of class wise performance, summarized in Table 3, our method performs best in terms of accuracy and F1-score on all classes except for class 3, which is blood collection. As shown in the confusion matrix in Figure 6, this class is mostly confused with blood glucose measurement. This is due to the fact that blood collection and blood glucose measurement are quite similar actions. CrossView fusion model performs best on class 12, which is diaper exchange and cleaning of area, reflecting this activity has highest variation from all other activities in the dataset. Overall, mean accuracy and F1 scores of class-wise performance for all classes, of our approach is 12.5\% better than the state-of-the-art KNN-based solution and the second-best ST-GCN-based solution.


\subsection{Ablation Studies}
\textbf{Single Modality vs. Dual Modality:} The impact of fusing acceleration and skeletal joints signal can be observed by comparing it with the single modality transformer models, trained on just skeletal joints or acceleration signal. We can see in Table 4, both spatio-temporal skeleton model and acceleration model give lower validation accuracy than CrossView fusion model. CrossView Fusion model out performs spatio-temporal skeleton model by 5\%.  However, the simple fusion method does not perform as good as single modality spatio-temporal skeleton model, which reflects that simple concatenation of the acceleration and skeletal joints feature vector, hurts the performance of model and makes it perform 1.7\% lower than skeleton only model. Among single modality models, we can see that spatio-temporal skeleton model performs 31\% better than acceleration model, and that makes sense because of the wide difference of sensor sampling rates (skeletal data was recorded at 100Hz and acceleration at just 4Hz, Figure 5), and more noise in acceleration data than skeletal joints data.
\begin{figure}
\centering
    \includegraphics[width=6cm]{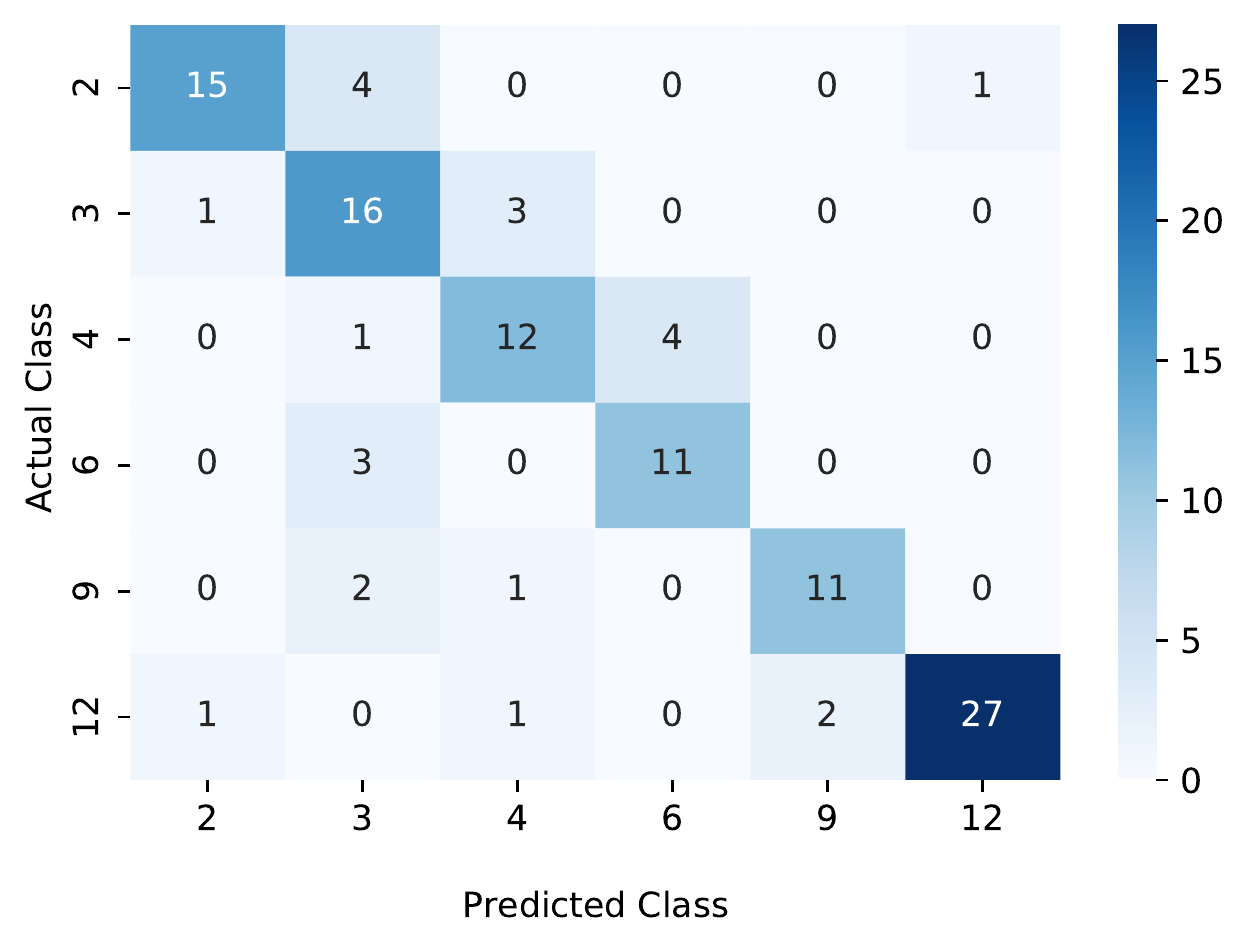} %
    \caption{\textbf{Confusion Matrix} of CrossView Fusion Model on Validation set. Activity2: Vital Signs Measurement, Activity3: Blood collection, Activity4: Blood glucose measurement, Activity6: Indwelling drip retention and connection, Activity 9: Oral care, Activity 12: Diaper exchange and cleaning of area.}%
    \label{fig:example}%
\end{figure}

\begin{figure*}[h]
\centering
    \includegraphics[width=17cm]{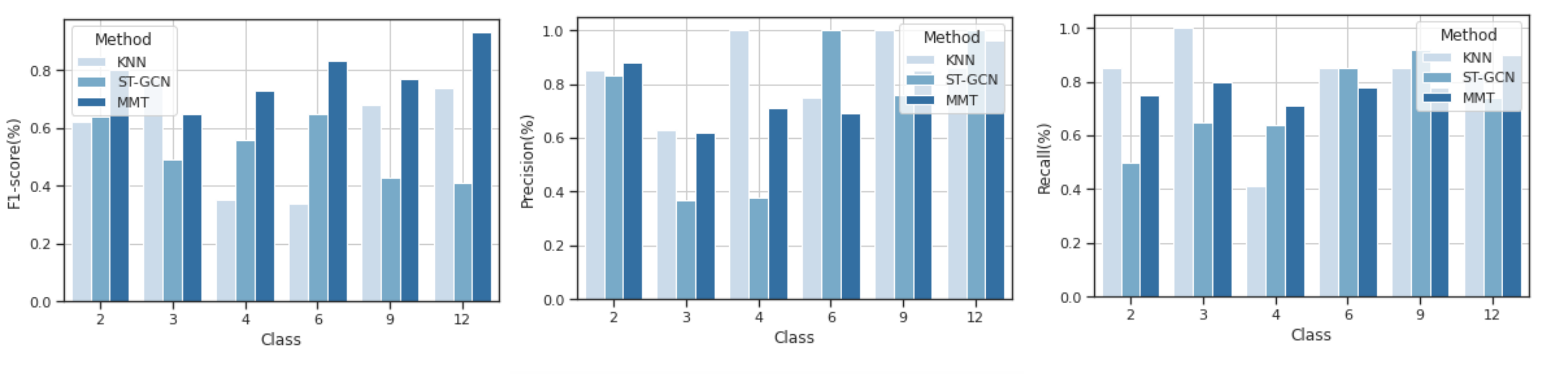} %
\caption{Class Wise F1-score, Precision and Recall comparison of \textbf{our Multi-Modal Transformer(MMT)} with top two solutions, \textbf{ST-GCN} and \textbf{KNN}. Our model MMT, outperforms ST-GCN and KNN in terms of F1-score for all classes and gives comparable performance in terms of precision and recall scores. }%
    \label{fig:example}%
\end{figure*}

\begin{figure}
\centering
    \includegraphics[width=6.5cm]{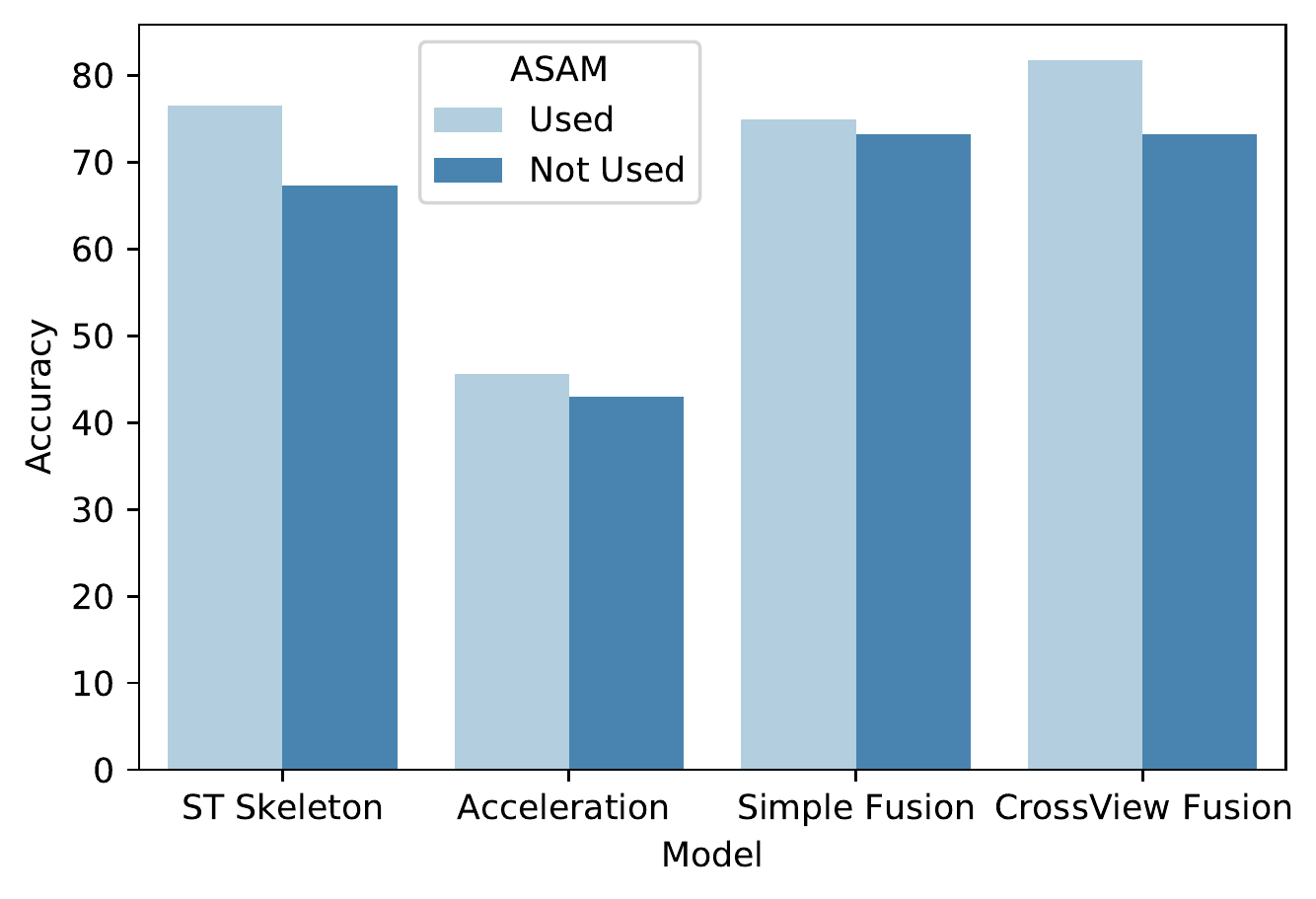} %
    \vspace{-0.5cm}
    \caption{Impact of using ASAM}%
    \vspace{-0.5cm}
\end{figure}

\textbf{Impact of CrossView Fusion:} We analyze the impact of CrossView Fusion mechanism in our multi-modal transformer by comparing it with the Simple Fusion variant. In Simple fusion model, we simply concatenate the CLS tokens coming from the single modality transformers, whereas in CrossView fusion model, we add cross attention based fusion between acceleration encoder and temporal encoder of spatio-temporal skeleton model. The impact of this added fusion can be seen in Table 4, where the CrossView fusion-based model outperforms the simple fusion model by 6.8\%. We can also see the features extracted by simple fusion method are different and not as diverse as CrossView fusion features, in Figure 4.

\textbf{Leave-One-Subject-Out Cross Validation}: For testing the generalizability of our proposed approach we perform leave-one-subject-out cross validation on the dataset. We combine test and train subjects and treat each subject as a test subject, while others as training subjects. Resultant performance of CrossView Fusion model for all subjects is reported in Table 5. Our proposed solution performed best for cross validation on subject 6, reflecting all other subjects made up a good diverse data for our solution to converge well and achieve 93.4\% accuracy. The worst validation scores were obtained for subject 7, for which the model only gave 61.1\% accuracy. Overall, we can see that the model is generalizing well and giving adequate performance on cross subject validation.

\textbf{Impact of using ASAM:}  We tried training all dual and single modality transformer models with and without ASAM\cite{Kwon2021ASAMAS}, as shown in Figure 8. Using ASAM allowed models to avoid over-fitting and generalize well. Single Modality models, Spatio-Temporal Skeleton model, and Acceleration model had less parameters so they have benefitted the least as compared to the CrossView fusion models which benefitted more from ASAM. The Acceleration model's accuracy improved by 2.6\% and skeleton model's accuracy improved by 9\% utilizing ASAM. We saw a boost of 2.7\% for simple fusion and a boost of 8.53\% for CrossView Fusion model by using ASAM. The CrossView fusion model had the largest gain from usage of ASAM, mainly because it has the highest number of parameters and a more bumpy loss function than all other models.

\section{Conclusion and Discussion}
\par In this work, we demonstrate the effectiveness of fusing acceleration and skeletal joints signals for performing skeletal action recognition. We present a novel multimodal transformer architecture with cross-attention-based fusion between skeletal joints and acceleration data. Our proposed multimodal fusion transformer model outperforms single modality and simple fusion baselines by a margin of 5-6\%. We achieve state-of-the-art results on the Nurse Care Activity Recognition dataset and illustrate generalizing ability of our method in ablation studies.

\textbf{Limitations and Future Work:}
\begin{itemize}
  \item  A limitation of the dataset includes highly imbalanced sampling rates of skeletal joints and acceleration signals, Figure 5, which proved to be an obstacle in unlocking the full potential of our proposed method. Exploring the impact of multimodal fusion transformers on a dataset with a uniform number of observations from acceleration and skeletal joints sensors might result in improved performance.
  \item The small size of the dataset is also a potential issue to resolve for future works.
  \item One can also explore pre-training the skeletal branch on skeletal joints data like NTU-RGB+D60/120 \cite{liu2020ntu,shahroudy2016ntu} and acceleration branch on NCRC-2 \cite{jem3-ap07-20} or NCRC-3 \cite{hj46-zs46-21} dataset to further improve the model's convergence. 
\end{itemize}

{\small
\bibliographystyle{ieee_fullname}
\bibliography{egbib}
}

\end{document}